\pdfoutput=1
\documentclass{article}
\usepackage[utf8]{inputenc}

\usepackage{amsmath,amssymb,amsthm}
\usepackage{amssymb,amsbsy}
\usepackage{mathtools}
\usepackage{bbm}
\usepackage{graphicx,xcolor}
\usepackage{subfigure}
\usepackage{cleveref}
\usepackage{url}
\usepackage{fancyvrb}
\usepackage{lipsum}
\usepackage[normalem]{ulem}

\newcommand\blfootnote[1]{%
  \begingroup
  \renewcommand\thefootnote{}\footnote{#1}%
  \addtocounter{footnote}{-1}%
  \endgroup
}

\theoremstyle{definition}

\theoremstyle{remark}

\def\tr{\mathop{\text{tr}}\kern.2ex}

\def\OA{\texttt{Interface}}

\usepackage{authblk}

\title{Arena: a toolkit for Multi-Agent Reinforcement Learning}

\author[1]{Qing Wang$^*$}
\author[1]{Jiechao Xiong$^*$}
\author[1]{Lei Han}
\author[1]{Meng Fang}
\author[1]{Xinghai Sun}
\author[2]{Zhuobin Zheng$^\dag$}
\author[1]{Peng Sun}
\author[1]{Zhengyou Zhang}
\affil[1]{Tencent AI Lab, Shenzhen}
\affil[2]{Tsinghua University}

\date{}

\begin{document}

\maketitle
\blfootnote{Preprint. Work in progress. 
($^*$) Equal contribution. ($^\dag$) Work was done during the internship with Tencent AI Lab.
}

\begin{abstract}
    We introduce Arena, a toolkit 
    for multi-agent reinforcement learning (MARL) research.
    In MARL, it usually requires customizing observations, rewards and actions for each agent, 
    changing cooperative-competitive agent-interaction, 
    and playing with/against a third-party agent, etc.
    We provide a novel modular design, called $\OA$, for manipulating such routines in essentially two ways:
    1) Different interfaces can be concatenated and combined,
    which extends the Open\-AI Gym Wrappers concept to MARL scenarios.
    2) During MARL training or testing, interfaces can be embedded in either wrapped
    OpenAI Gym compatible \emph{Environments} or raw environment compatible \emph{Agents}.
    We offer off-the-shelf interfaces for several popular MARL platforms, 
    including StarCraft II, Pommerman, ViZDoom, Soccer, etc.
    The interfaces effectively support self-play RL and cooperative-competitive hybrid MARL.
    Also, Arena can be conveniently extended to your own favorite MARL platform. 

\end{abstract}

\section{Introduction}
In recent years, deep reinforcement learning have achieved human-level performance on various challenging tasks \cite{mnih2015human, silver2016mastering, openai2018five, jaderberg2018human, deepmind2019alphastar}.
Similar to supervised learning, where data with finely tuned features and loss is the key of good performance, the success of reinforcement learning (RL) relies on high quality environments with well-designed observations, action spaces, and reward functions.
Different settings and combinations of observation transformation, action space definition, and reward reshaping are usually experimented in the engineering process.
Among various work on providing environments for RL \cite{greg2016openai,beattie2016deepmind,nichol2018gotta,tassa2018deepmind}, OpenAI Gym \cite{greg2016openai} {uses} the concept of ``wrapper'', to enable a flexible and modular way of organizing different state and action settings.
A Gym wrapper can change the observation, reward, and action space of an environment, and is the basic building block for constructing more complex environments.
However, OpenAI Gym considers mostly single agent case in its original design. In multi-agent reinforcement learning (MARL), we may encounter new problems like combining multiple cooperative agents as a team, and testing with different competitive agents, for which the Gym wrapper may be inadequate.

In this work, we introduce the Arena project which uses ``interface'' to represent observation and action transformation in a more modular way.
The Arena interface is mostly compatible with Gym wrappers and can be stacked similarly. In addition, these interfaces can be wrapped on both environments and agents, and support side by side combining. Thus it provides better flexibility for multi-agent training and testing. For the initial release, we provide supporting interfaces for 6 multi-agent game environments. The supported games range from simplest two-player Pong game, which can be solved with a single machine under half an hour, to the full game of StarCraft II, which may require days of training on a cluster of machines. We also provide performance of baseline methods for these environments. We hope Arena can contribute to the research of multi-agent reinforcement learning and game-theoretical learning community.

\section{Motivation}

\subsection{The loop of reinforcement learning}

A wide range of real-world problems can be formulated as sequential decision problems. These problems usually involve the interaction between an ``environment'' and one or more ``agents''. 
For the case that only one agent interacts with an environment, the problem is usually modeled as a Markov Decision Process (MDP) \cite{bellman1957markovian} or Partially Observable Markov Decision Process (POMDP) \cite{aastrom1965optimal}; while for the case that multiple agents interact with a single environment simultaneously, the problem is often formulated as a stochastic game \cite{shapley1953stochastic}.
In both models, the agent(s) observe the (full or partial) state of an environment, and decide what actions to take in each step. The environment then evolves depending on the actions, and provides new observations for the agents as the next loop. In the training phase, the environment also emits rewards as feedback to guide the learning of agents. The basic loop is depicted in Figure \ref{fig:gymwrapper}. The study of reinforcement learning is about achieving maximum (discounted accumulated or average) rewards.

\subsection{Observation and action transformation: the Gym way}

\begin{figure}[ht]
    \centering
    \includegraphics[clip, trim=0cm 15.5cm 20cm 0cm, width=0.7\textwidth]{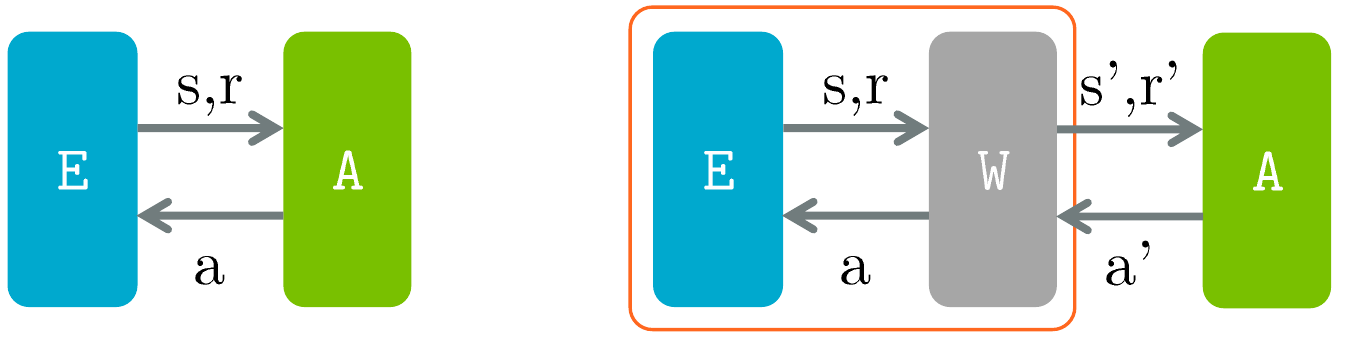}
    \caption{\textbf{Left}: A standard loop in reinforcement learning: agents receive observation (and reward) from the environment; And the environment receives agents actions then evolves to next state. \textbf{Right}: A gym wrapper can change the observation, reward, and action space of the wrapped environment, thus is convenient for training learn-able agents with structured inputs and outputs.}
    \label{fig:gymwrapper}
\end{figure}

To represent the decision function (policy) of an agent, the raw observation of the environment is usually pre-processed into structured ``features'', then feed to a trainable model (e.g.~neural networks) to get structured outputs, and post-processed into raw actions for the environment. Different choices of modeling and algorithms may require different observations and actions. To support this, OpenAI Gym \cite{greg2016openai} introduced the concept of \verb|Wrapper|, which can be stacked and nested, to construct environments with different observations and actions. The concept is illustrated at the right side of Figure \ref{fig:gymwrapper}. The Gym environments and its flexible and convenient \verb|Wrapper|s are widely used in the community of reinforcement learning research.

\subsection{The problem of multiple agents}

The MDP model commonly used in reinforcement learning {considers} a single agent only. However, for more general problems, it is common that several agents interact with each other in a single environment. For example, in many team-based tasks, several agents act as a team to achieve a common goal. These fully cooperative multi-agent problems are usually modeled as a stochastic game with a universal reward function shared among all agents \cite{foerster2018counterfactual, han2019grid}. On the other hand, in many competitive games, we are also interested in finding the equilibrium of agents' strategy. These problems are naturally modeled as a stochastic game with agent-specific reward functions. And in more complicated cases, there is cooperation within a team and competition between teams at the same time. We may want to share information (agent's observation) within a team and act as a whole (collaboratively) from other teams' perspective. For these problems, we would like to construct an environment with structured observations and actions for training homogeneous agents, as well as to set up a game between heterogeneous agents trained with different observations and action spaces for evaluation. We find that wrapping all observation and action transformations on the environment may not be adequate for these needs.

\subsection{A solution for multi-agent: The Arena Interface}

To deal with these problems, 
we try to provide more flexible ``wrapper''s for multi-agent training and testing. 
Therefore, we introduce the Arena \verb|Interface|. The concept of \verb|Interface| is like \verb|Wrapper| in OpenAI Gym. In each multi-agent environment, the observations for agents are wrapped in a tuple, as well as rewards and actions, which is similar to the Gym-like design in \cite{mordatch2017emergence,bansal2017emergent}. With each \verb|Interface|, we can transform the observation and action from input or previous interface to the next one.
The \verb|Interface| is designed with 4 rules:
\begin{itemize}
    \item \verb|Interface|s can be stacked and combined to form a new \verb|Interface|.
    \item \verb|Interface|s can be wrapped on both environment and agent.
    \item \verb|Interface|s can be wrapped on multiple agents to form a team.
    \item \verb|Interface|s should be compatible with Gym \verb|Wrapper|s when wrapped on an environment.
\end{itemize}
We will elaborate on these points in the following paragraphs. A rigorous definition (class template) of \verb|Interface| can be found in the source code.

\subsubsection{Stacking and Combining}
\begin{figure}[ht]
    \centering
    \includegraphics[clip, trim=0cm 15cm 24cm 0cm, width=0.5\textwidth]{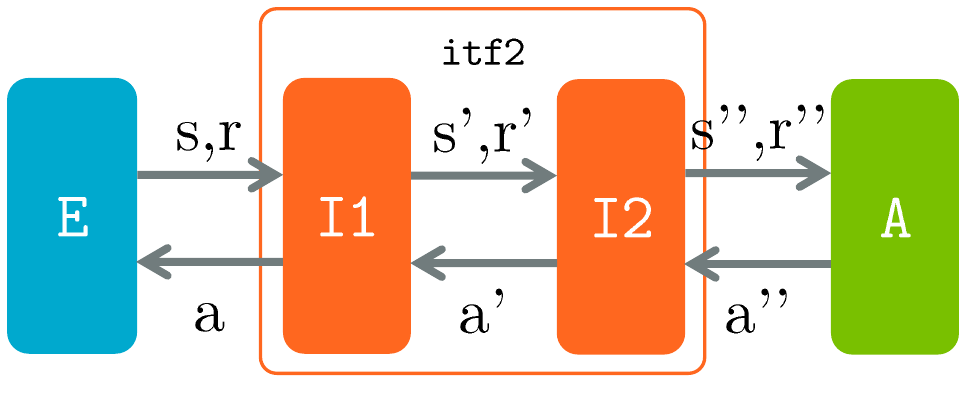}
    \caption{Interface Stacking: In this example, the interface I2 is stacked over I1. The observation from environment is firstly processed by I1 then by I2. On the contrary, the action from agent is firstly processed by I2 then by I1.}
    \label{fig:1}
\end{figure}

Similar to Gym wrappers, \verb|Interface|s can be stacked together simply as
\begin{Verbatim}[frame=single]
 itf1 = I1(None)
 itf2 = I2(itf1)
\end{Verbatim}
which stacked a new interface \verb|I2| over \verb|I1|. In the environment-agent loop, the observation from environment is firstly processed by \verb|I1|, and the result is further processed by \verb|I2|. The action outputted by agent is firstly processed by \verb|I2|, and then by \verb|I1|. The RL loop with these interfaces is depicted in Figure \ref{fig:1}.

\begin{figure}[ht]
    \centering
    \includegraphics[clip, trim=0cm 14.8cm 24cm 0cm, width=0.5\textwidth]{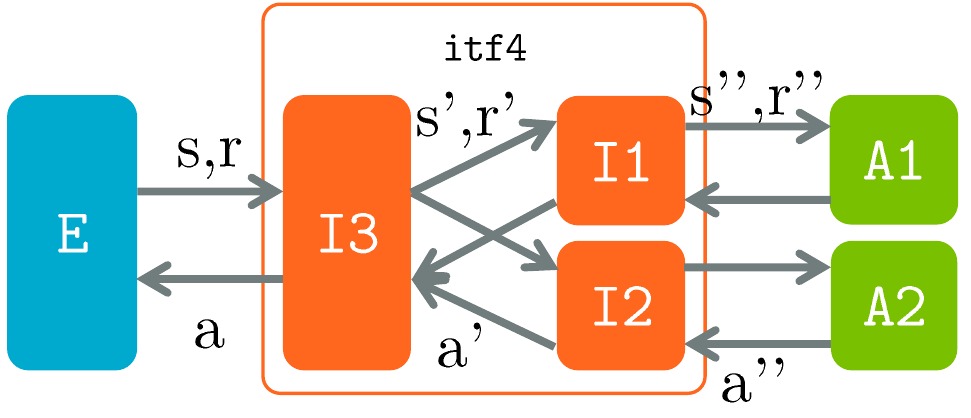}
    \caption{Interface Combination: The interfaces I1 and I2 are combined and stacked over I3. The observation transformed by I3 is split to I1 and I2; while the actions processed by I1 and I2 are combined before processed by I3.}
    \label{fig:intercombine}
\end{figure}

We also provide mechanism for combining multiple parallel interfaces for multi-agent environments. This is especially useful for sharing information among fully-cooperative agents. Consider the interfaces defined as following:
\begin{Verbatim}[frame=single]
 itf1 = I1(None)
 itf2 = I2(None)
 itf3 = I3(None)
 itf4 = Combine(itf3, [itf1,itf2])
\end{Verbatim}
The two interfaces \verb|I1| and \verb|I2| are combined side by side, and then stacked with interface \verb|I3|. The relational structure of these interfaces is depicted in Figure \ref{fig:intercombine}.

\subsubsection{Wrapping on both environment and agent}
\begin{figure}[ht]
    \centering
    \includegraphics[clip, trim=0cm 15.5cm 16.5cm 0cm, width=0.9\textwidth]{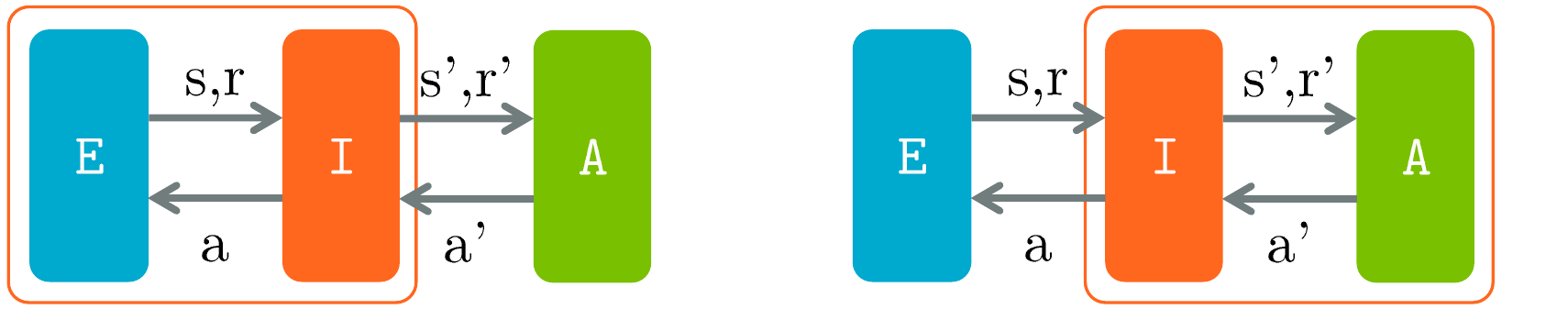}
    \caption{In Arena, we support wrapping an interface on an environment (left) as well as on an agent (right). The flexibility of wrapping interfaces enables a general solution for evaluating agents trained in varied ways.}
    \label{fig:envagtint}
\end{figure}

For the ease of training and testing, we provide functionalities to wrap an interface on an environment, as well as on an agent. With \verb|EnvItfWrapper|, we can wrap an interface on an environment, forming a new environment with observation transformation and action transformation implicitly called in \verb|env.step()|. The resulting environment can have structured observation and action space, which is convenient for the training of a learn-able agent. On the other hand, we can also wrap an interface on an agent with \verb|AgtItfWrapper|. By hiding the observation transformation and action transformation in \verb|agent.step()|, the resulting agents all accept the observation in the original observation space, and return actions in the original action space. As a result, we can support battles between agents with different observation spaces and action spaces with \verb|Interface| wrapped on these agents. The concept is illustrated in Figure \ref{fig:envagtint}.

\begin{figure}[ht]
    \centering
    \includegraphics[clip, trim=0cm 15cm 26cm 0cm, width=0.4\textwidth]{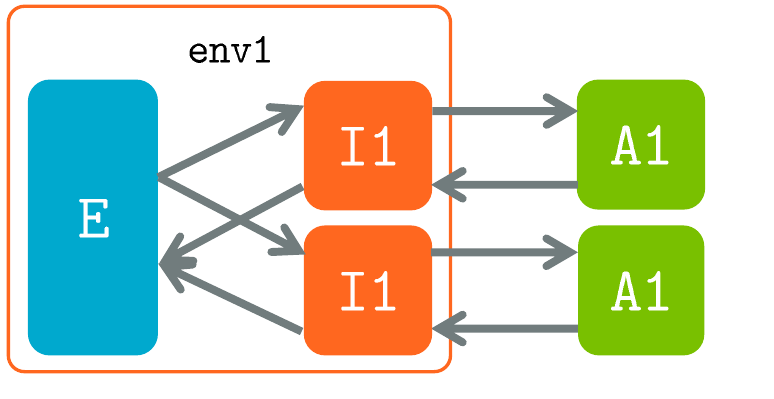}
    \hspace{3em}
    \includegraphics[clip, trim=0cm 15cm 26cm 0cm, width=0.4\textwidth]{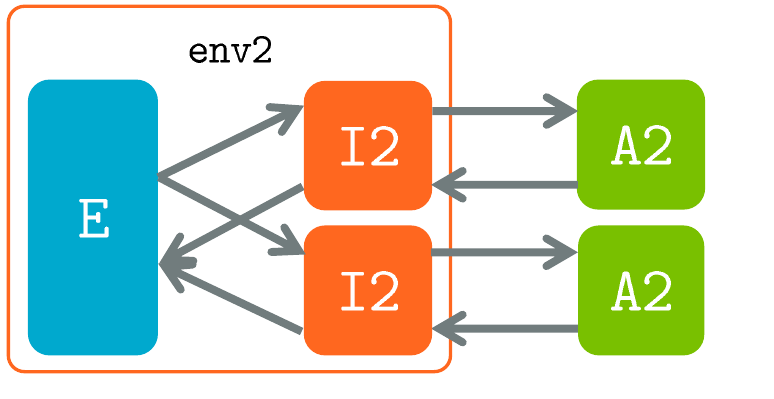}
    \includegraphics[clip, trim=0cm 14.5cm 26cm 0cm, width=0.4\textwidth]{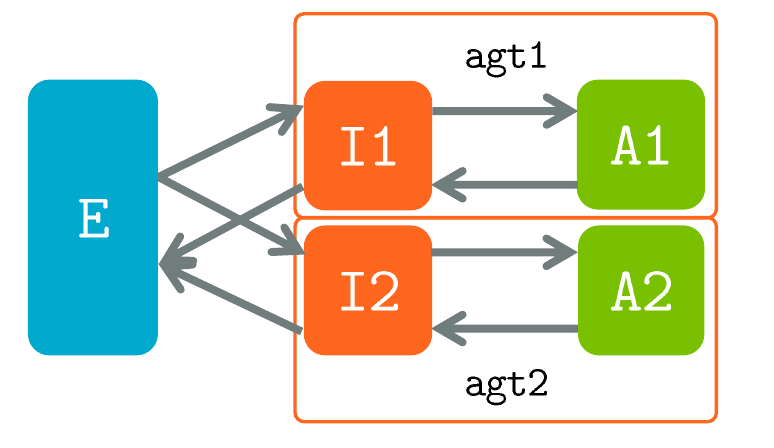}
    \caption{\textbf{Upper}: For training homogeneous agents, we can wrap an interface to the original multi-agent environment. The observations transformed by the interface are received by each agent respectively, and agents' actions are transformed by the interface before received by the original environment.
    \textbf{Lower}: For testing heterogeneous agents, the interface used by each agent in the training phase is wrapped on itself respectively. The wrapped agents then receive the original observations from the multi-agent environment, and output actions that conform to the original environment's action space.}
    \label{fig:multi}
\end{figure}

For example, we can train agents with two different interfaces, and evaluate them against each other with the following codes. The relational diagram for these codes is shown in Figure \ref{fig:multi}.
\begin{Verbatim}[frame=single]
 env1   = EnvItfWrapper(Env(), [I1(None), I1(None)])
 model1 = Train(env1)
 env2   = EnvItfWrapper(Env(), [I2(None), I2(None)])
 model2 = Train(env2)
 agt1   = AgtItfWrapper(model1, I1(None))
 agt2   = AgtItfWrapper(model2, I2(None))
 Test(env, [agt1, agt2])
\end{Verbatim}

\subsubsection{Combining multiple agents as a team}
\begin{figure}[ht]
    \centering
    \includegraphics[clip, trim=0cm 15cm 26cm 0cm, width=0.42\textwidth]{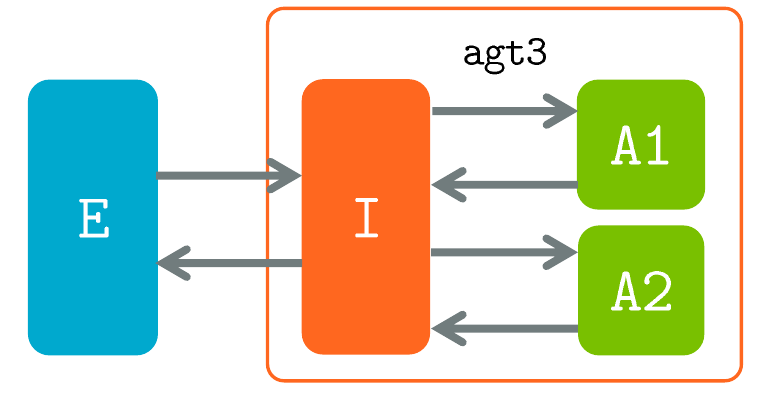}
    \caption{Combining agents as a team: The combined team receives a tuple of observations and outputs a tuple of actions for the basic 2 agents.}
    \label{fig:agtintwrapper}
\end{figure}

To have better support for multi-agent environments, we also provide {a team mechanism to group agents}. A ``team'' shares the same function calls as an ``agent''. It can be seen as a generalized agent which can actually control multiple original agents. In the following example, we will define a team of two agents, with the \verb|AgtItfWrapper|. The resulting team \verb|agt3| is the combination of two agents instantiated by \verb|A1| and \verb|A2|, respectively. An illustrative graph for the team structure can be found in Figure \ref{fig:agtintwrapper}.
\begin{Verbatim}[frame=single]
 itf  = I()
 agt3 = AgtItfWrapper([A1(), A2()], itf)
\end{Verbatim}

\subsubsection{Compatible with Gym Wrappers}
The Arena \verb|Interface| is compatible with OpenAI Gym \verb|Wrappers|. In Arena, an environment inherits property fields and methods from \verb|Env| in OpenAI Gym. With \verb|EnvItfWrapper| aforementioned, the \verb|Interface|s can be used interchangeably with \verb|Wrappers| on environments. In addition, we also standardize the properties and methods for an agent, which is not defined in OpenAI Gym. The agent methods are designed such that the RL loop can be formed with a Gym environment and an Arena agent. With these designs, we hope that Arena interfaces can be best compatible with existing projects based on OpenAI Gym.
\begin{Verbatim}[frame=single]
 env  = Env()
 env1 = MyWrapper(env)
 env2 = EnvItfWrapper(env1, MyItf())
\end{Verbatim}
\begin{figure}[ht]
    \centering
    \includegraphics[clip, trim=0cm 14.2cm 23cm 0cm, width=0.55\textwidth]{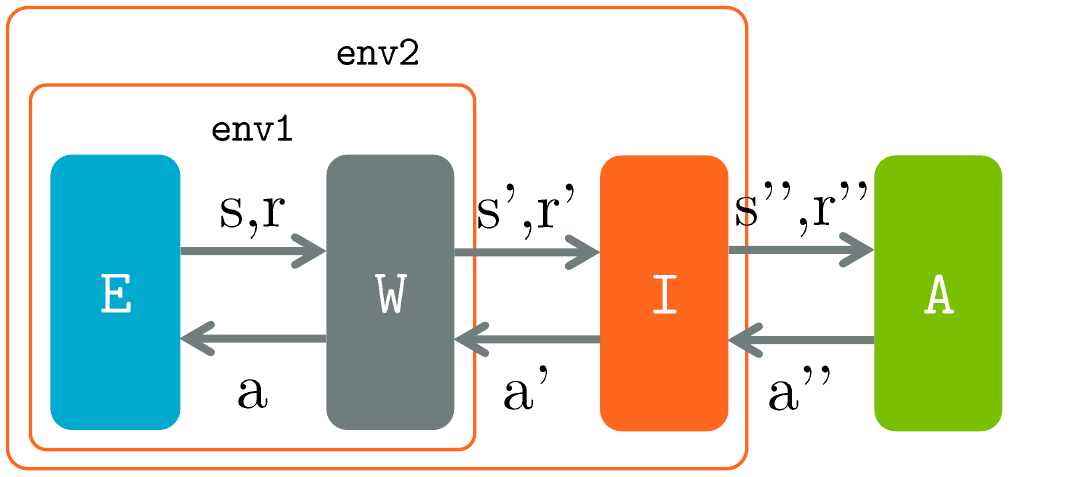}
    \caption{Gym Compatibility: In this example, we firstly wrap a gym wrapper over the environment, then wrap an interface over it. The reverse order of wrapping also works although the resulting environment may differ.}
    \label{fig:gymcompatible}
\end{figure}

\section{Related Work}

The research of reinforcement learning is fueled by popular and easy-to-use environment packages, e.g.~OpenAI Gym \cite{greg2016openai} and Deepmind Lab \cite{beattie2016deepmind}. OpenAI Gym \cite{greg2016openai} includes a collection of environments for classical reinforcement learning, and is widely used by the research community. Our work can be viewed as an extension of Gym for multi-agent cases. And Deepmind Lab \cite{beattie2016deepmind} is mainly focused on first-person 3D games, while also providing simple and flexible API.

For multi-agent reinforcement learning (MARL), the work of \cite{mordatch2017emergence} provides particle environments for both cooperative and competitive settings, as well as the training method and algorithm performances for these environments.
Also, in the work of \cite{bansal2017emergent}, the authors show that in a simple competitive multi-agent environment, the trained agents can have very complex behaviors.
The environment for their experiments is also open-sourced.
For related work on StarCraft, a suite of multi-agent environments has been proposed and open-sourced in the work of \cite{samvelyan19smac}, together with performance baselines of several training methods.

\section{Supported Environments}

For the initial release, we provide interfaces together with 6 environments, namely \textit{Pong-2p}, \textit{SC2Battle}, \textit{SC2Full}, \textit{PommerMan}, \textit{ViZDoom}, and \textit{Soccer}. These games are selected as they range from simplest two-player Pong game, to complex full-length StarCraft II game. \verb|Interface|s compatible with these environments are provided as examples for the practical usage of Arena Interfaces on multi-agent environments. In addition, we also provide experimental performance of baseline RL methods on these environments. We leave detailed description of these interfaces and baseline performance in Appendix for interested readers.
\begin{figure}[ht]
    \centering
    \includegraphics[height=0.23\textwidth,width=0.32\textwidth]{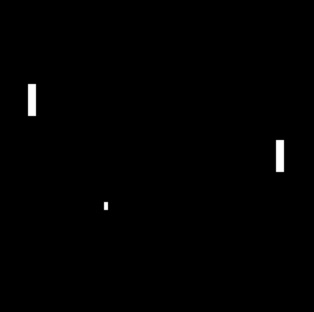}
    \includegraphics[height=0.23\textwidth,width=0.32\textwidth]{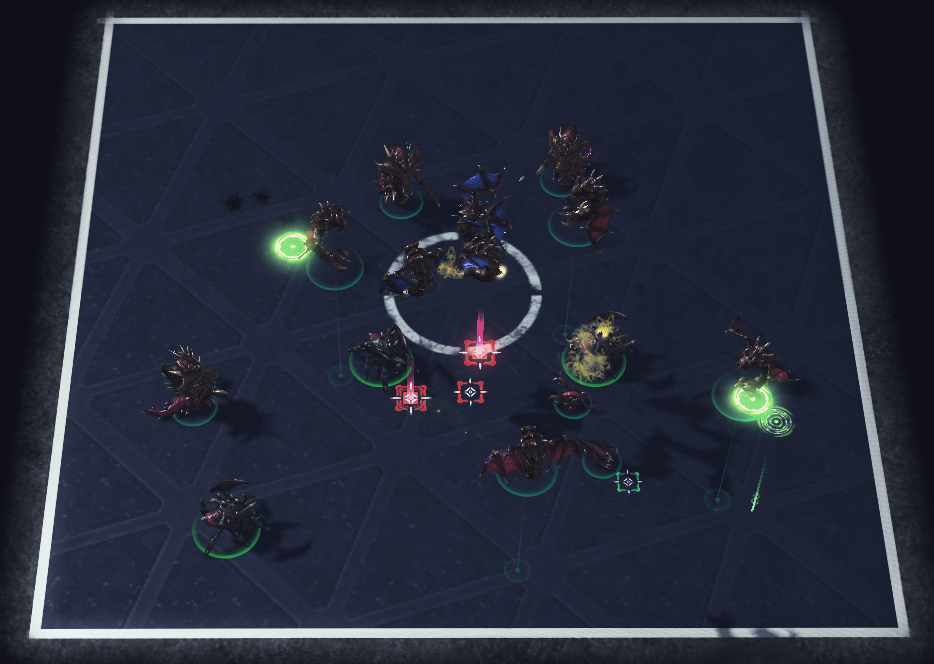}
    \includegraphics[height=0.23\textwidth,width=0.32\textwidth]{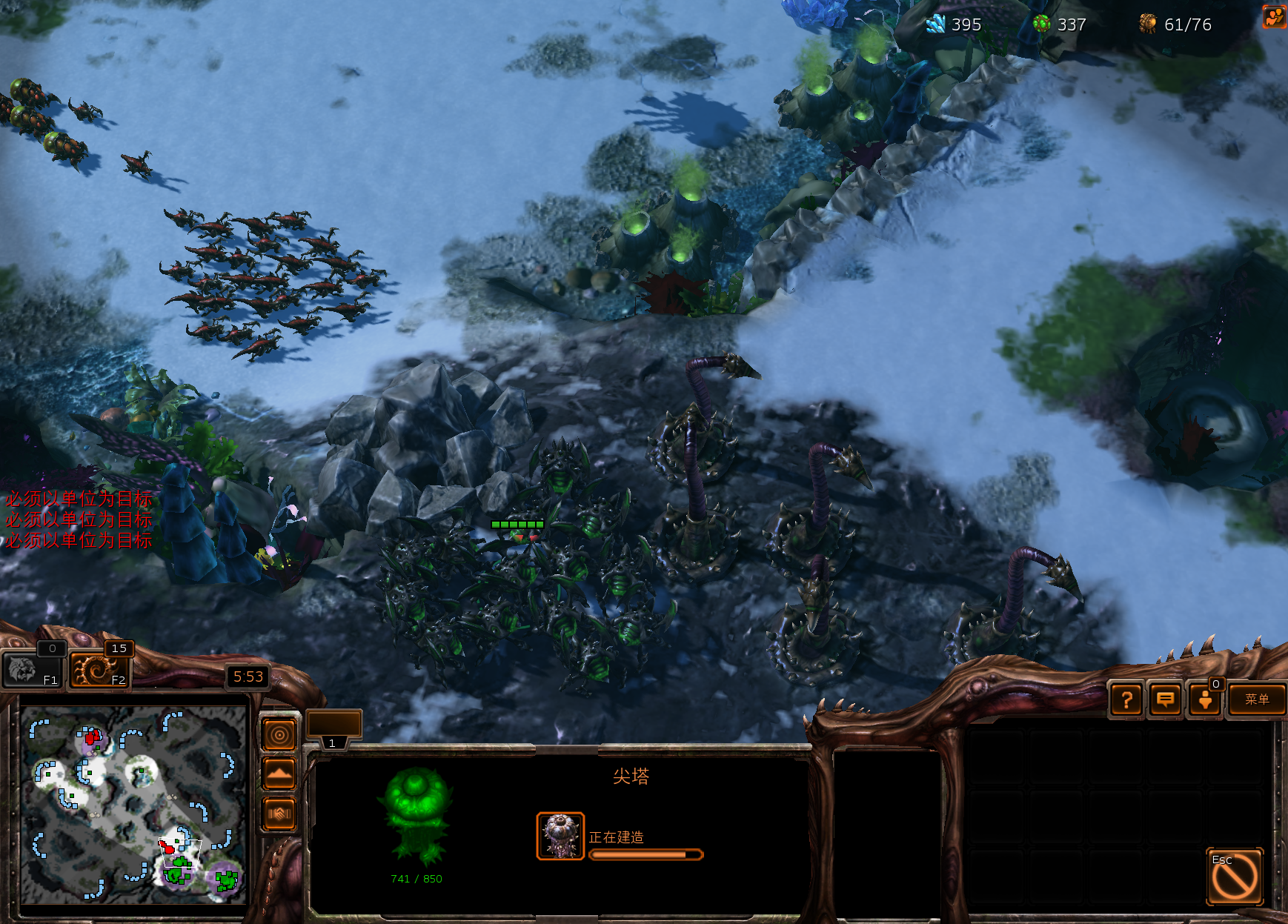}
    \includegraphics[height=0.23\textwidth,width=0.32\textwidth]{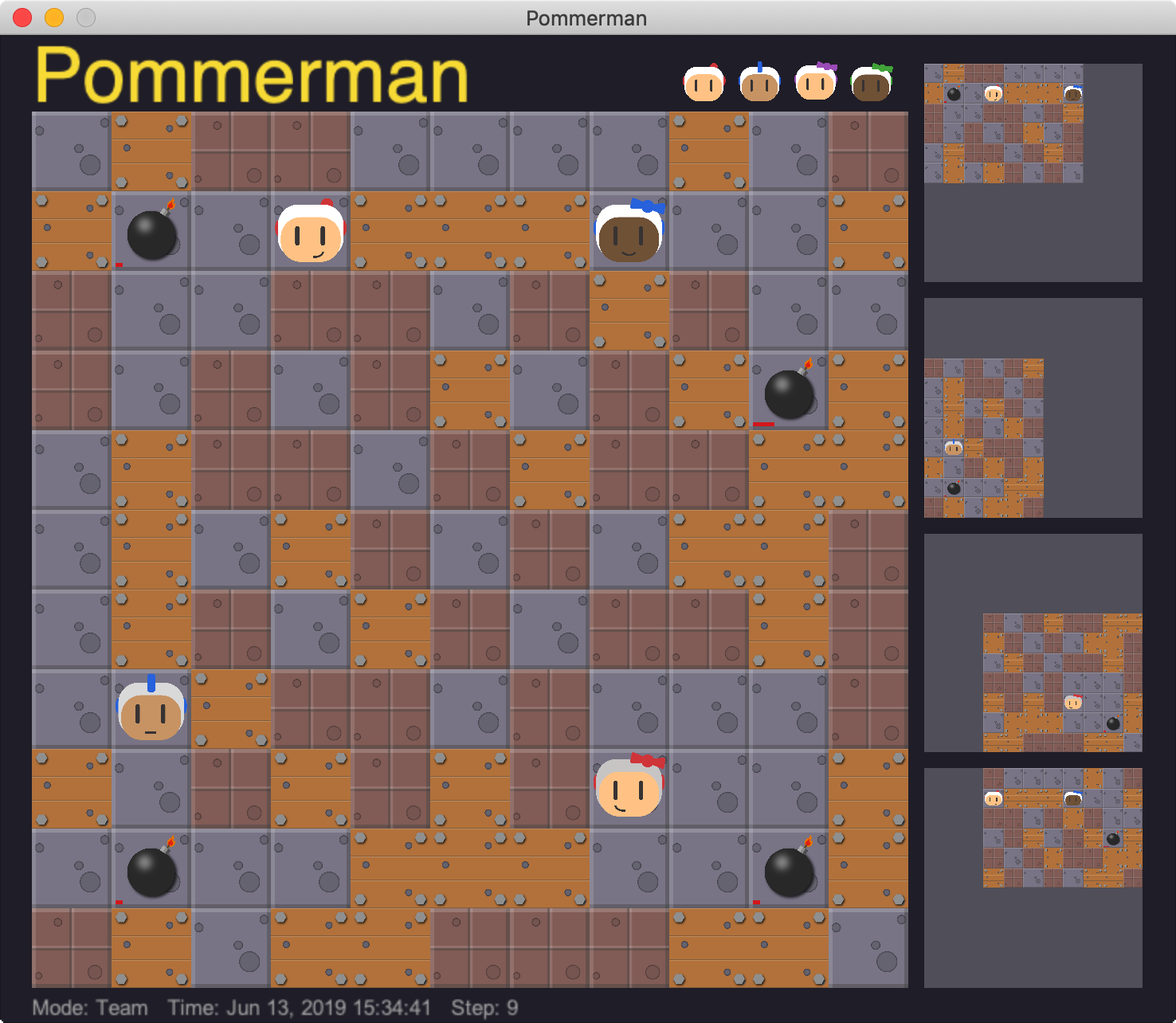}
    \includegraphics[height=0.23\textwidth,width=0.32\textwidth]{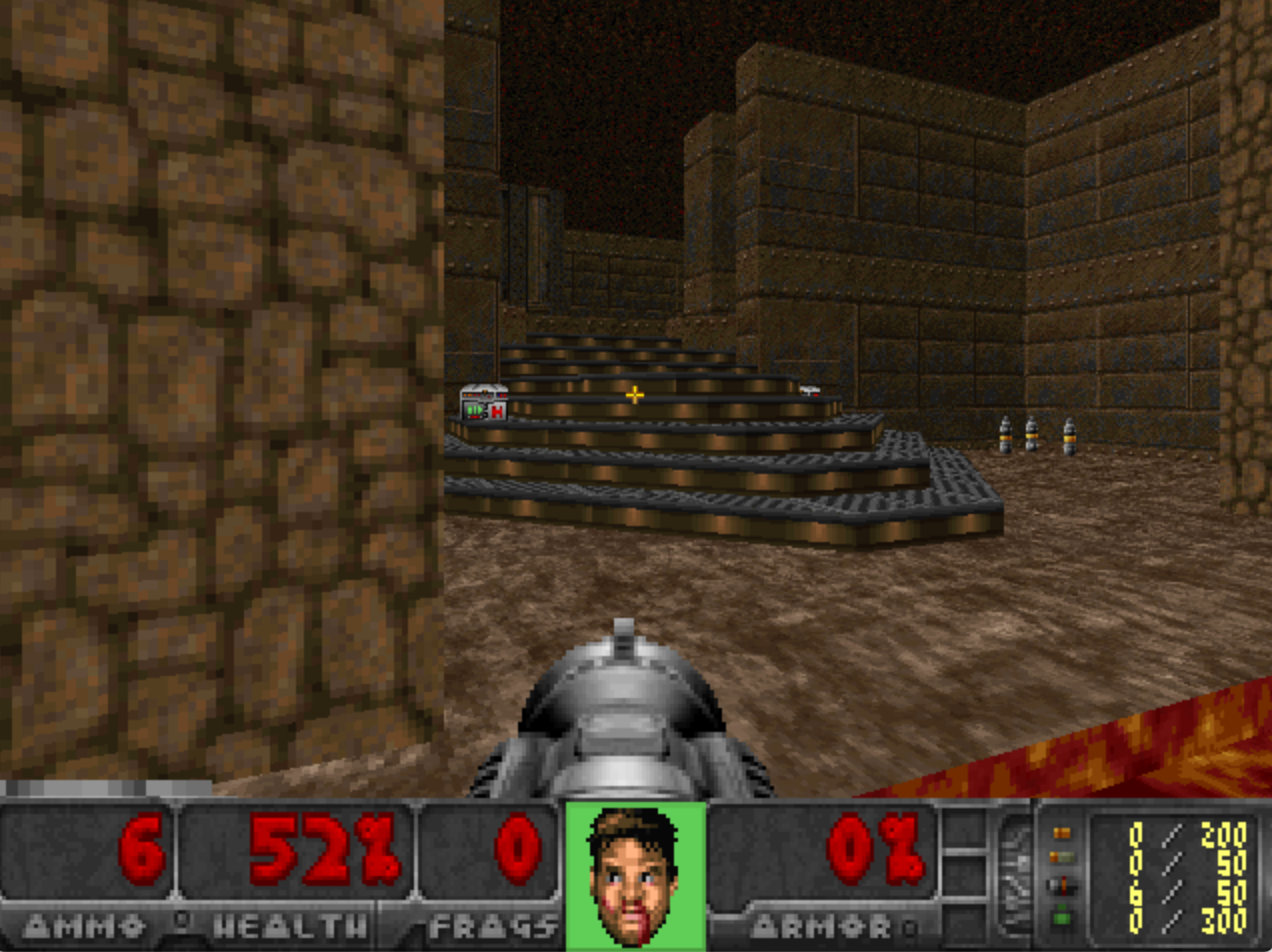}
    \includegraphics[height=0.23\textwidth,width=0.32\textwidth]{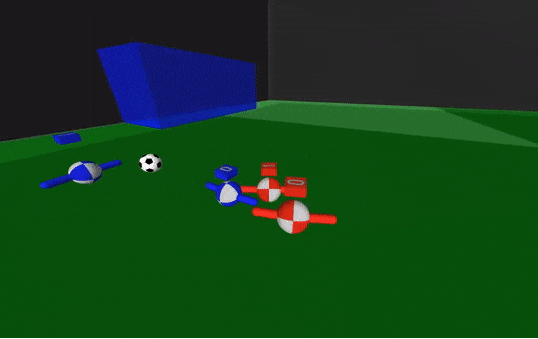}
    \caption{Screenshots of currently supported environments in Arena.
    \textbf{Upper} (from left to right): Pong-2p, SC2Battle, and SC2Full (based on \cite{vinyals2017starcraft}).
    \textbf{Lower} (from left to right): PommerMan \cite{resnick2018pommerman}, ViZDoom \cite{wydmuch2018vizdoom}, and Soccer \cite{liu2019emergent}.
    }
    \label{fig:screenshot}
    \vspace{-2em}
\end{figure}

\section{Conclusion and Future Direction}
In this work we propose Arena, a new toolkit for multi-agent reinforcement learning research. We provide modular components for implementing diversified multi-agent environments and agents with a universal interface. The framework is validated with experiments with popular multi-agent baseline methods. In the future, we are working on a general training and evaluation platform for cooperative and competitive MARL.

\section*{Acknowledgement}
Thanks Jiaming Lu for his work on Pong-2p and ViZDoom. Thanks Yali Du for her work in SC2 battle games. Thanks Jing Li, Liheng Chen, Yingru Li for many insightful discussions. Thanks Wenhan Luo, Yuan Gao for their early efforts in ViZDoom competition. Thanks Rongjun Qin for his early work in ViZDoom. Thanks Shiyu Huang and Jia Xu for the helpful discussion on ViZDoom MARL settings. Thanks Ji Liu for the inspiring discussions. Thanks Han Liu and Tong Zhang for organizing and involving the early work. 

\bibliographystyle{unsrt}
\bibliography{main}

\begin{thebibliography}{10}

\bibitem{mnih2015human}
Volodymyr Mnih, Koray Kavukcuoglu, David Silver, Andrei~A Rusu, Joel Veness,
  Marc~G Bellemare, Alex Graves, Martin Riedmiller, Andreas~K Fidjeland, Georg
  Ostrovski, Stig Petersen, Charles Beattie, Amir Sadik, Ioannis Antonoglou,
  Helen King, Dharshan Kumaran, Daan Wierstra, Shane Legg, and Demis Hassabis.
\newblock Human-level control through deep reinforcement learning.
\newblock {\em Nature}, 518(7540):529--533, 2015.

\bibitem{silver2016mastering}
David Silver, Aja Huang, Chris~J Maddison, Arthur Guez, Laurent Sifre, George
  Van Den~Driessche, Julian Schrittwieser, Ioannis Antonoglou, Veda
  Panneershelvam, Marc Lanctot, Sander Dieleman, Dominik Grewe, John Nham, Nal
  Kalchbrenner, Ilya Sutskever, Timothy Lillicrap, Madeleine Leach, Koray
  Kavukcuoglu, Thore Graepel, and Demis Hassabis.
\newblock Mastering the game of go with deep neural networks and tree search.
\newblock {\em Nature}, 529(7587):484--489, 2016.

\bibitem{openai2018five}
OpenAI.
\newblock Openai five.
\newblock \url{https://blog.openai.com/openai-five/}, 2018.

\bibitem{jaderberg2018human}
Max Jaderberg, Wojciech~M. Czarnecki, Iain Dunning, Luke Marris, Guy Lever,
  Antonio~Garc{\'{\i}}a Casta{\~{n}}eda, Charles Beattie, Neil~C. Rabinowitz,
  Ari~S. Morcos, Avraham Ruderman, Nicolas Sonnerat, Tim Green, Louise Deason,
  Joel~Z. Leibo, David Silver, Demis Hassabis, Koray Kavukcuoglu, and Thore
  Graepel.
\newblock Human-level performance in first-person multiplayer games with
  population-based deep reinforcement learning.
\newblock {\em CoRR}, abs/1807.01281, 2018.

\bibitem{deepmind2019alphastar}
The~AlphaStar team.
\newblock Alphastar: Mastering the real-time strategy game starcraft ii.
\newblock
  \url{https://deepmind.com/blog/alphastar-mastering-real-time-strategy-game-starcraft-ii/},
  2019.

\bibitem{greg2016openai}
Greg Brockman, Vicki Cheung, Ludwig Pettersson, Jonas Schneider, John Schulman,
  Jie Tang, and Wojciech Zaremba.
\newblock Openai gym, 2016.

\bibitem{beattie2016deepmind}
Charles Beattie, Joel~Z Leibo, Denis Teplyashin, Tom Ward, Marcus Wainwright,
  Heinrich K{\"u}ttler, Andrew Lefrancq, Simon Green, V{\'\i}ctor Vald{\'e}s,
  Amir Sadik, et~al.
\newblock Deepmind lab.
\newblock {\em arXiv preprint arXiv:1612.03801}, 2016.

\bibitem{nichol2018gotta}
Alex Nichol, Vicki Pfau, Christopher Hesse, Oleg Klimov, and John Schulman.
\newblock Gotta learn fast: A new benchmark for generalization in rl.
\newblock {\em arXiv preprint arXiv:1804.03720}, 2018.

\bibitem{tassa2018deepmind}
Yuval Tassa, Yotam Doron, Alistair Muldal, Tom Erez, Yazhe Li, Diego de~Las
  Casas, David Budden, Abbas Abdolmaleki, Josh Merel, Andrew Lefrancq, et~al.
\newblock Deepmind control suite.
\newblock {\em arXiv preprint arXiv:1801.00690}, 2018.

\bibitem{bellman1957markovian}
Richard Bellman.
\newblock A markovian decision process.
\newblock {\em Journal of Mathematics and Mechanics}, pages 679--684, 1957.

\bibitem{aastrom1965optimal}
Karl~Johan {\AA}str{\"o}m.
\newblock Optimal control of markov processes with incomplete state
  information.
\newblock {\em Journal of Mathematical Analysis and Applications},
  10(1):174--205, 1965.

\bibitem{shapley1953stochastic}
Lloyd~S Shapley.
\newblock Stochastic games.
\newblock {\em Proceedings of the national academy of sciences},
  39(10):1095--1100, 1953.

\bibitem{foerster2018counterfactual}
Jakob~N Foerster, Gregory Farquhar, Triantafyllos Afouras, Nantas Nardelli, and
  Shimon Whiteson.
\newblock Counterfactual multi-agent policy gradients.
\newblock In {\em Thirty-Second AAAI Conference on Artificial Intelligence},
  2018.

\bibitem{han2019grid}
Lei Han, Peng Sun, Yali Du, Jiechao Xiong, Qing Wang, Xinghai Sun, Han Liu, and
  Tong Zhang.
\newblock Grid-wise control for multi-agent reinforcement learning in video
  game {AI}.
\newblock In {\em Proceedings of the 36th International Conference on Machine
  Learning (ICML)}, pages 2576--2585, 2019.

\bibitem{mordatch2017emergence}
Igor Mordatch and Pieter Abbeel.
\newblock Emergence of grounded compositional language in multi-agent
  populations.
\newblock {\em arXiv preprint arXiv:1703.04908}, 2017.

\bibitem{bansal2017emergent}
Trapit Bansal, Jakub Pachocki, Szymon Sidor, Ilya Sutskever, and Igor Mordatch.
\newblock Emergent complexity via multi-agent competition.
\newblock {\em arXiv preprint arXiv:1710.03748}, 2017.

\bibitem{samvelyan19smac}
Mikayel Samvelyan, Tabish Rashid, Christian~Schroeder de~Witt, Gregory
  Farquhar, Nantas Nardelli, Tim G.~J. Rudner, Chia-Man Hung, Philiph H.~S.
  Torr, Jakob Foerster, and Shimon Whiteson.
\newblock {The} {StarCraft} {Multi}-{Agent} {Challenge}.
\newblock {\em CoRR}, abs/1902.04043, 2019.

\bibitem{vinyals2017starcraft}
Oriol Vinyals, Timo Ewalds, Sergey Bartunov, Petko Georgiev, Alexander~Sasha
  Vezhnevets, Michelle Yeo, Alireza Makhzani, Heinrich K{\"{u}}ttler, John
  Agapiou, Julian Schrittwieser, John Quan, Stephen Gaffney, Stig Petersen,
  Karen Simonyan, Tom Schaul, Hado van Hasselt, David Silver, Timothy~P.
  Lillicrap, Kevin Calderone, Paul Keet, Anthony Brunasso, David Lawrence,
  Anders Ekermo, Jacob Repp, and Rodney Tsing.
\newblock Starcraft ii: A new challenge for reinforcement learning.
\newblock {\em arXiv preprint arXiv:1708.04782}, 2017.

\bibitem{resnick2018pommerman}
Cinjon Resnick, Wes Eldridge, David Ha, Denny Britz, Jakob Foerster, Julian
  Togelius, Kyunghyun Cho, and Joan Bruna.
\newblock Pommerman: A multi-agent playground.
\newblock {\em arXiv preprint arXiv:1809.07124}, 2018.

\bibitem{wydmuch2018vizdoom}
Marek Wydmuch, Micha{\l} Kempka, and Wojciech Ja{\'s}kowski.
\newblock Vizdoom competitions: Playing doom from pixels.
\newblock {\em IEEE Transactions on Games}, 2018.

\bibitem{liu2019emergent}
Siqi Liu, Guy Lever, Josh Merel, Saran Tunyasuvunakool, Nicolas Heess, and
  Thore Graepel.
\newblock Emergent coordination through competition.
\newblock {\em arXiv preprint arXiv:1902.07151}, 2019.

\bibitem{xinghaipong}
Xinghai Sun.
\newblock Xinghai sun pong.
\newblock
  \url{https://github.com/xinghai-sun/deep-rl/blob/master/docs/selfplay_pong.md}.

\bibitem{stevenpong}
Steven Hewitt.
\newblock Steven hewitt pong.
\newblock \url{https://github.com/Steven-Hewitt/Multi-Agent-Pong-Rally}.

\bibitem{sun2018tstarbots}
Peng Sun, Xinghai Sun, Lei Han, Jiechao Xiong, Qing Wang, Bo~Li, Yang Zheng,
  Ji~Liu, Yongsheng Liu, Han Liu, and Tong Zhang.
\newblock Tstarbots: Defeating the cheating level builtin ai in starcraft ii in
  the full game.
\newblock {\em arXiv preprint arXiv:1809.07193}, 2018.

\bibitem{Kempka2016ViZDoom}
Micha{\l} Kempka, Marek Wydmuch, Grzegorz Runc, Jakub Toczek, and Wojciech
  Ja\'skowski.
\newblock {ViZDoom}: A {D}oom-based {AI} research platform for visual
  reinforcement learning.
\newblock In {\em IEEE Conference on Computational Intelligence and Games},
  pages 341--348, Santorini, Greece, Sep 2016. IEEE.
\newblock The best paper award.

\bibitem{wu2016training}
Yuxin Wu and Yuandong Tian.
\newblock Training agent for first-person shooter game with actor-critic
  curriculum learning.
\newblock In {\em ICLR}, 2016.

\bibitem{dosovitskiy2016learning}
Alexey Dosovitskiy and Vladlen Koltun.
\newblock Learning to act by predicting the future.
\newblock In {\em ICLR}, 2017.

\bibitem{huang2019combo}
Shiyu Huang, Hang Su, Jun Zhu, and Ting Chen.
\newblock Combo-action: Training agent for fps game with auxiliary tasks.
\newblock In {\em AAAI}, 2019.

\bibitem{todorov2012mujoco}
Emanuel Todorov, Tom Erez, and Yuval Tassa.
\newblock Mujoco: A physics engine for model-based control.
\newblock In {\em Intelligent Robots and Systems (IROS), 2012 IEEE/RSJ
  International Conference on}, pages 5026--5033. IEEE, 2012.

\end{thebibliography}

\newpage

\appendix

\section{List of supported environments}

\subsection{Pong-2p}
Pong-2p (Pong of 2 players) is much like the Atari Pong~\cite{greg2016openai}, 
except that the two players on both sides are controllable.
Due to its simplicity, Pong-2p {usually} serves as a ``sanity-check'' environment
in the sense that one can quickly {train} a strong policy with any decent competitive MARL algorithm.

Our code is inspired by other 2-player Pong implementations, e.g.,~\cite{xinghaipong,stevenpong}. However, we carefully adjust and modify the physics and dynamics in regards of the ball's speed, the reflection and collision with the pad in different positions, etc.,
so as to ensure that a reasonable policy is learn-able.
For the baseline, 
we also {provide} a rule-based built-in AI which naively follows the ball in vertical direction. 

\begin{figure}[h]
    \centering
    \includegraphics[width=0.6\textwidth]{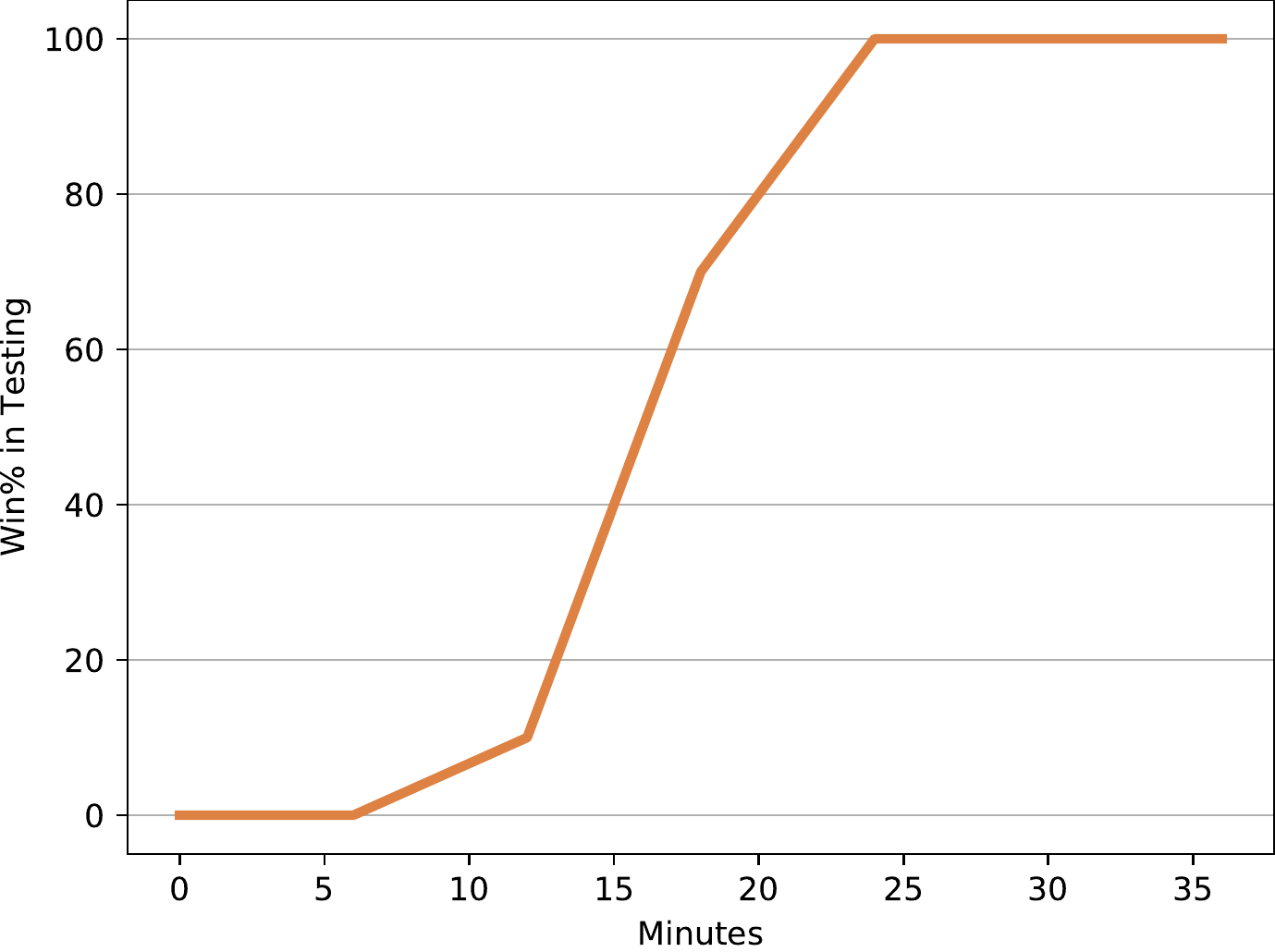}
    \caption{Win-rate against rule-based Agent in Pong-2p. Horizontal: self-play training hours; Vertical: Win-rate in separate testing. }
    \label{fig:pong-result}
\end{figure}

Besides the raw environment, we additionally provide Arena interfaces that expose the observation in 2D image and the action in discrete action space. 
We then employ a simple self-play algorithm {similar to} \cite{bansal2017emergent,openai2018five},
where we use 1 Learner (GPU) and 30 Actors (CPUs) in a single machine.
We found that a ConvNet based neural network model quickly defeats the built-in AI unseen during training,
as shown in Fig.~\ref{fig:pong-result}.

\subsection{SC2Battle}
We study many challenging battle games in StarCraft II based on the learning environment SC2LE \cite{vinyals2017starcraft}. StarCraft II is a famous real-time strategy game with annual career league among professional human players. In SC2Battle, we are mostly interested in training multiple agents to collaborate as a team to defeat the other team. We provide some symmetric battle scenarios that have been studied in \cite{han2019grid} as below: 
\begin{description}
\item[5I] 5 Immortals vs. 5 Immortals, where Immortal is a range unit of Protoss that can attack enemies at a distance;
\item[3I2Z] (3 Immortals, 2 Zealots) vs. (3 Immortals, 2 Zealots), where Zealot is a melee unit of Protoss that attacks enemies standing close to it;
\item[MAB] Mixed army battle, in which the battle is taken place among a random number of mixed units of Zerg, containing Baneling, Zergling, Roach, Hydralisk and Mutalisk. These units will be explained in details in the supplementary material.
\end{description}

In \cite{han2019grid}, the observations and the actions are organized into some image-like structures and we provide the detailed interfaces encoding these observation and action spaces in the Arena platform. Specifically, the following observation and action interfaces are provided for the SC2Battle scenarios.
\subsubsection{Observation Interfaces}
\begin{description}
\item[Img5IObsItf] is the observation interface used in 5I. The observations are represented as a grid feature map with size (8, 8, 6) with 6 channels including the health points (HP), shield and cooling down (CD) for the units in both sides. 
\item[Img3I2ZObsItf] is the observation interface used in 3I2Z. The size of the grid feature map is (8, 8, 16) with 16 channels including the HP, shield, CD and damage for both Immortal and Zealot on both sides.
\item[ImgMABObsItf] is the observation interface used in MAB. The size of the grid feature map is (16, 16, 18) with 18 channels including Mutalisk HP, Hydralisk HP, Roach HP, Zergling HP, Baneling HP, CD, attack range, damage and whether it can attack air units for both sides. The number of units from each unit type will be randomized over episodes in the range [1, 5]. On MAB, we use a larger feature map with size 16$\times$16, since there might be much more units in this setting.
\end{description}

\subsubsection{Action Interfaces}
The action interfaces for SC2 battle games are provided as below.
\begin{description}
\item[ImgActItf] encodes the action space into an image-like structure. The action space of each agent in 5I and 3I2Z contains 9 discrete actions with 8 move directions and the action of attacking the nearest enemy. 
\end{description}
\begin{description}
\item[ImgMABActItf] is the interface used for MAB. The action space of the agents in MAB contains 10 discrete actions with 8 move directions, attacking the nearest ground unit in priority, and attacking the nearest air unit in priority. It is worth mentioning that in MAB, the Mutalisk is an air unit and it does not collide with other units, so on MAB several units could overlap in the same grid. So, we distinguish the air units and the ground units in channels. That is, we use different channels in action maps for the ground and air units. 
\end{description}

\subsubsection{Experiment}
\begin{figure*}[t]
\centering
\subfigure[5I]{
\includegraphics[width=0.3\linewidth]{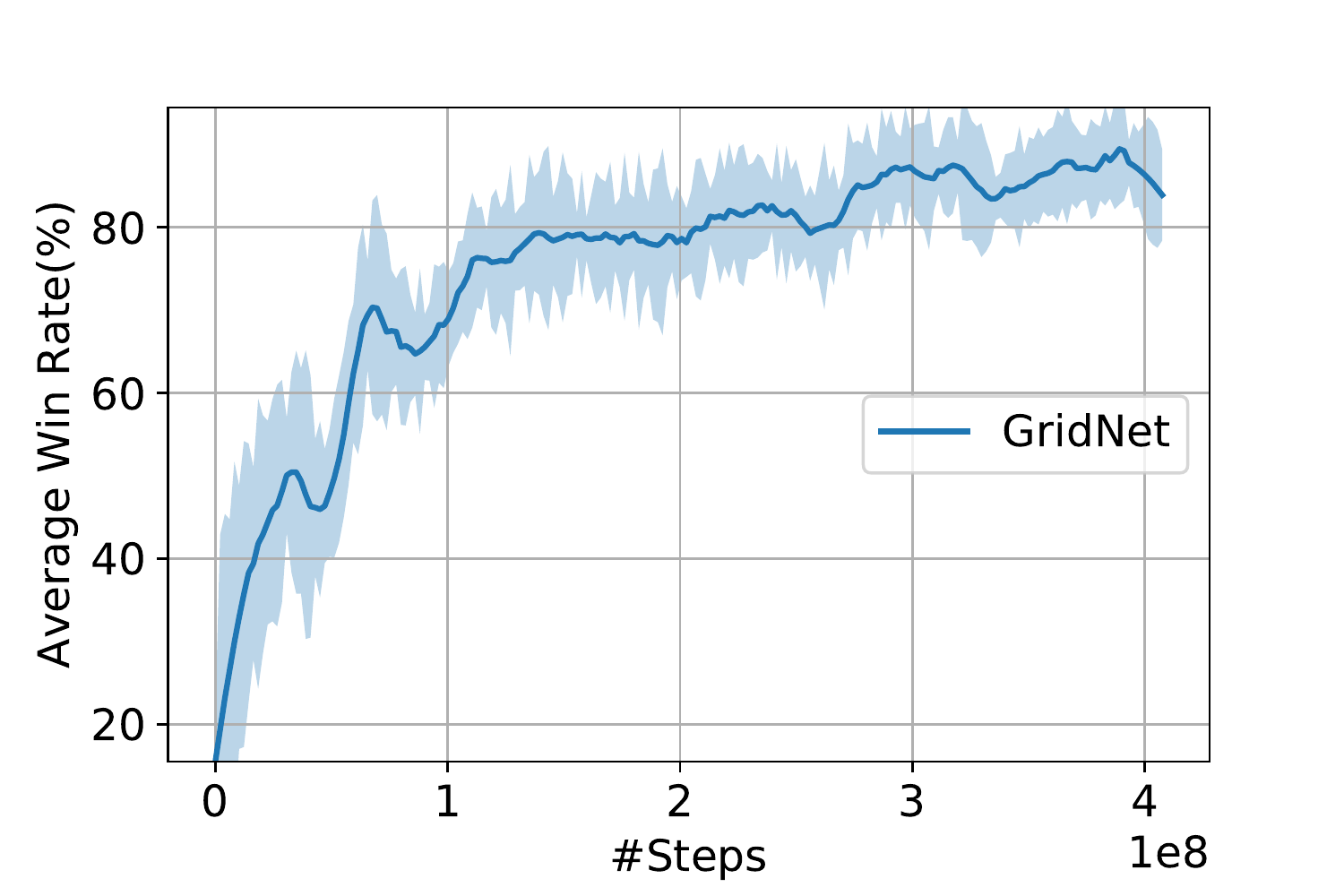}
}
\subfigure[3I2Z]{
\includegraphics[width=0.3\linewidth]{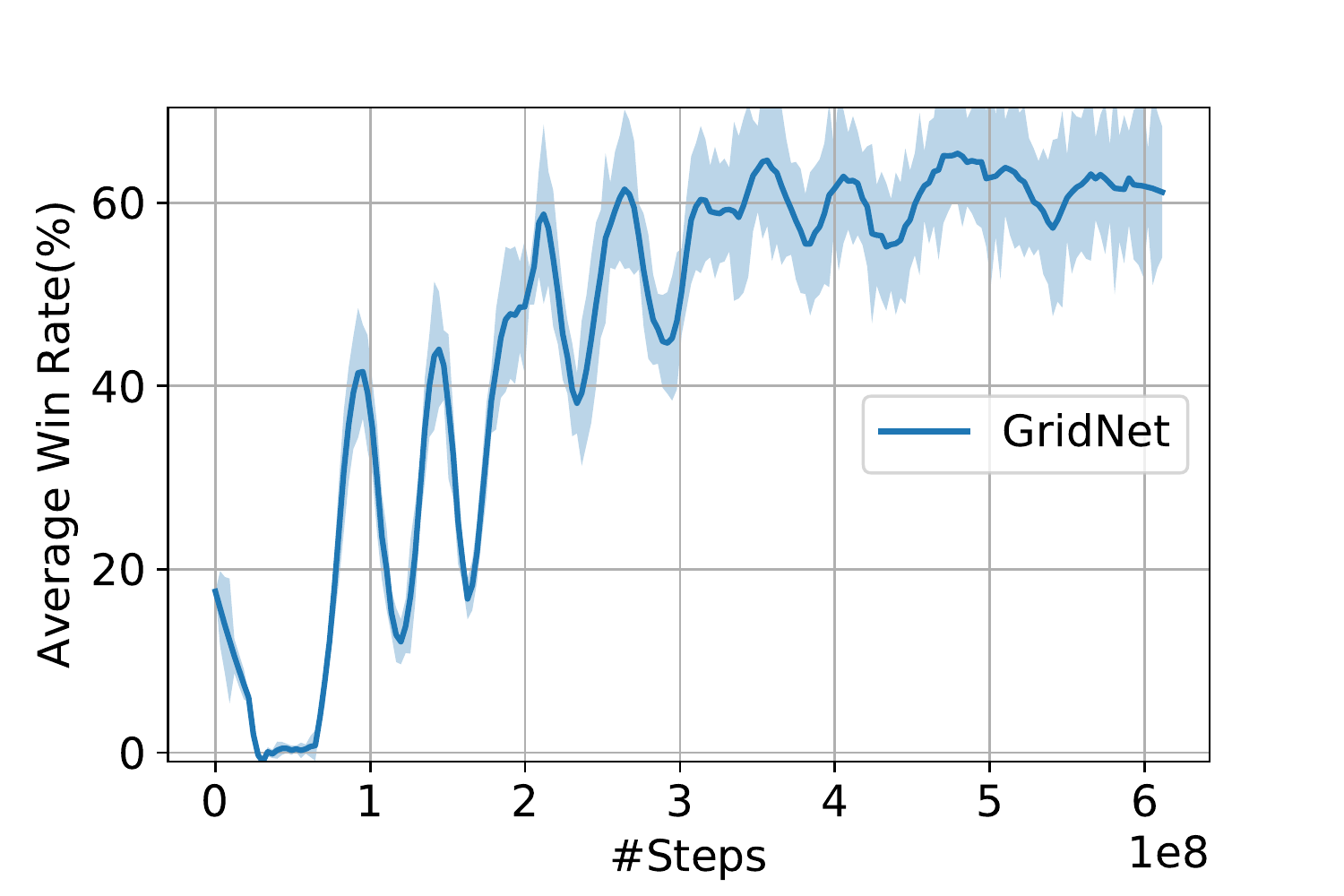}
}
\subfigure[MAB]{
\includegraphics[width=0.3\linewidth]{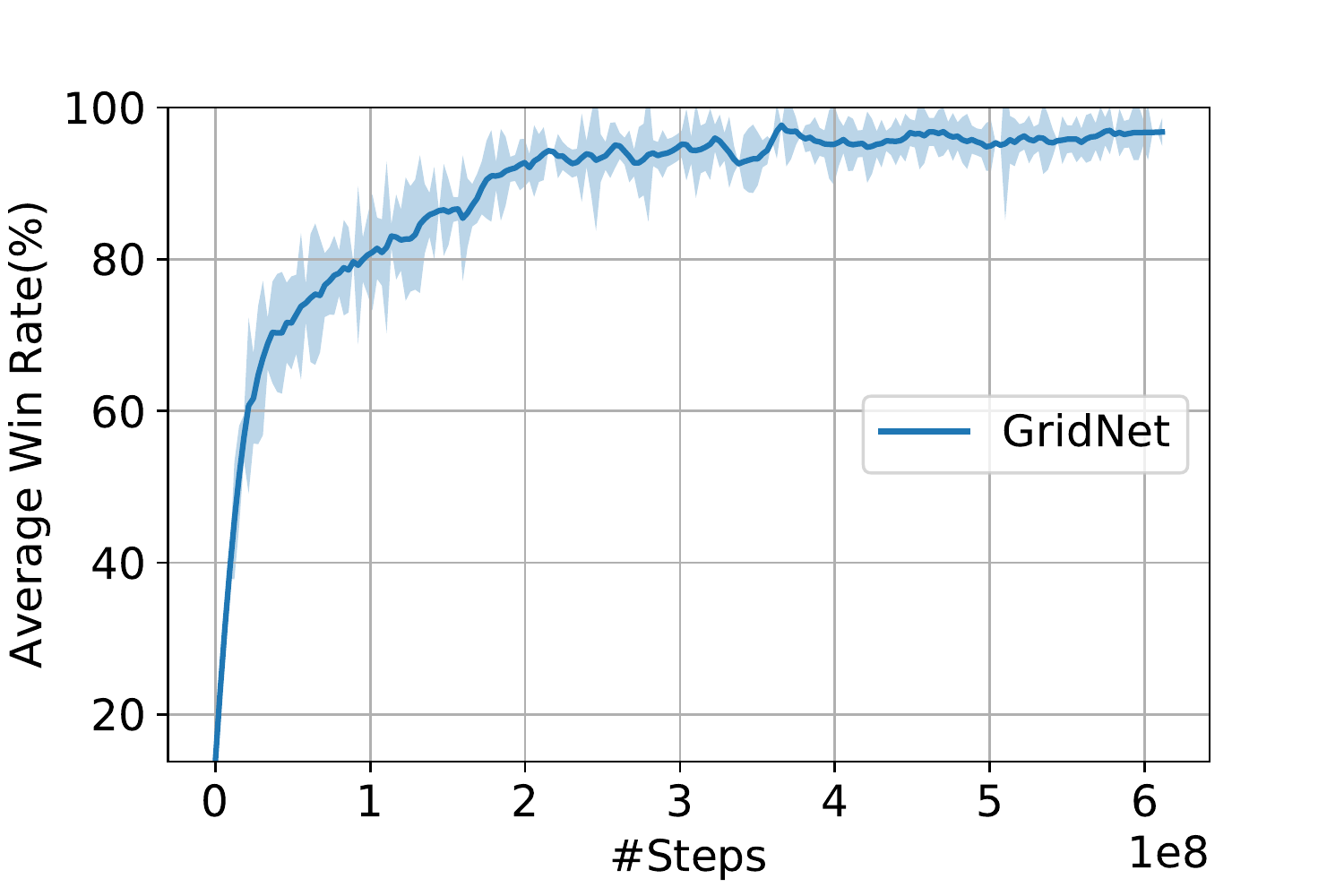}
}
\caption{Average test wining rate vs.~training steps (i.e., number of passed data points).}
\label{fig:wr}
\end{figure*}
We provide some benchmark results based on the SC2 battle scenarios described above. More detailed experimental results conducted on these games can be found in \cite{han2019grid}. 

For all the 3 settings, we randomly initialize the agent status, including health points, shield, cooling down and locations in the training phase to provide sufficient diversity. For testing, we remove most of the randomness and only keep randomized locations. 
For battle games, e.g., any settings mentioned above, units might die once the combat takes place, and the number of agents will decrease. We will append fake observations with some additional features to indicate whether an agent is dead or not.

We use the GridNet method proposed in \cite{han2019grid}. We train the agents with self-play, i.e., training the policies with their historical models.
We store the trained policies every 100 training batches and evaluate each stored policy over 100 testing episodes to report the average test winning rate. A benchmark result against a rule-based hit-and-run policy is given in Fig.~\ref{fig:wr}. As we can observe, the test winning rate keeps increasing as the training proceeds and it achieves faster convergence.

\subsection{SC2Full}

{StarCraft II full game is widely considered as the most challenging Real Time Strategy (RTS) game, due to its large observation space and continuous (infinite) action space. Among various work on modeling agents, TStarBots \cite{sun2018tstarbots} provide two agents for 1v1 Zerg-vs-Zerg full game, which can beat cheating-level built-in AIs.} Specifically, TstarBot1 is based on deep reinforcement learning over flat actions, and TstarBot2 is based on rule controllers over hierarchical actions. Based on these two agents, Arena provides observation interfaces as used in TstarBot1 and action interfaces of two bots for research convenience. {We will document these interfaces briefly in the following sections.}
\subsubsection{Observation Interfaces}
\begin{description}
\item[ZergSC2LearnerObsItf] This is the observation interface used in TstarBot1. It includes both spatial features and non-spatial features. Spatial features are multi-channel images of allied (and rival) unit counts, where different channels represent different kinds of unit types. Non-spatial features include \emph{UnitStatCountFeature} (allied/rival all/combat/ground/air units count), \emph{UnitTypeCountFeature} (units count of each type), \emph{PlayerFeature} (food, mineral, gas etc.), \emph{GameProgressFeature} (one-hot features of game loop), \emph{WorkerFeature} (mineral/gas/idle workers count), \emph{ScoreFeature} (scores provided by game core).
\item[ZergAreaObsItf] This interface appends to observation a vector including the features of all resource area, allied/rival units count of each unit type, base hp ratio, mineral/gas worker vacancy, idle worker number, amount of mineral resource left and gas resource left of each area.
\item[ZergTechObsItf] This interface appends to observation a vector representing our technology upgrades.
\item[ZergUnitProgObsItf] This interface appends to observation a vector representing our building and technology research progress. Each kind of building/tech has two features, a 0/1 indicating whether the building/tech is in progress, and a float number ranging from 0 to 1 indicating the actual progress of building/researching.
\end{description}

\subsubsection{Action Interfaces}
Arena provides the following action interfaces for StarCraft II full game based on two TstarBots.

\textbf{ZergFlatActItf:} This action interface is based on the action space of TstarBot1. It provides a flat action space with 165 macro actions described in Table 1 in \cite{sun2018tstarbots}.

The action space in TstarBot2 is organized in a two-tier structure with both macro actions and micro actions. Here we fix the lower level rule-based micro action controller and abstract the high level macro based controller as interfaces. The main advantage of such action space is that the controllers at the same tier can take simultaneous actions. Each controller can be easily replaced by the rule-based controller in TstarBot2.
\begin{description}
\item[ZergDataItf] Interface to set up \emph{Data Context \& Action Manager} of TstarBot2. This interface provides basic facilities for the following interfaces.
\item[ZergProdActItf] Action interface based on the rule-based \emph{Production Manager} in TstarBot2. Each action represents producing a specific type of unit, building a building, or researching a tech.
\item[ZergResourceActItf] Action interface based on the rule-based \emph{Resource Manager} in TstarBot2. Each action is a macro action, including ``gas-first'' (higher priority for harvesting gas), ``mineral-first'' (higher priority for harvesting mineral), and ``none''(repeating last strategy). \emph{Resource Manager} will translate the high level strategy to detailed micro actions.
\item[ZergCombatActItf] Action interface based on the rule-based \emph{Combat Manager} in TstarBot2. Each action is a macro action, including ``attack specific resource area'', ``attack nearest enemy'', ``defend'', ``rally around own base'', and ``rally around rival base'' with all combat units.
\end{description}

\subsubsection{Experiment}
\begin{figure}[ht]
    \centering
    \includegraphics[width=0.7\textwidth]{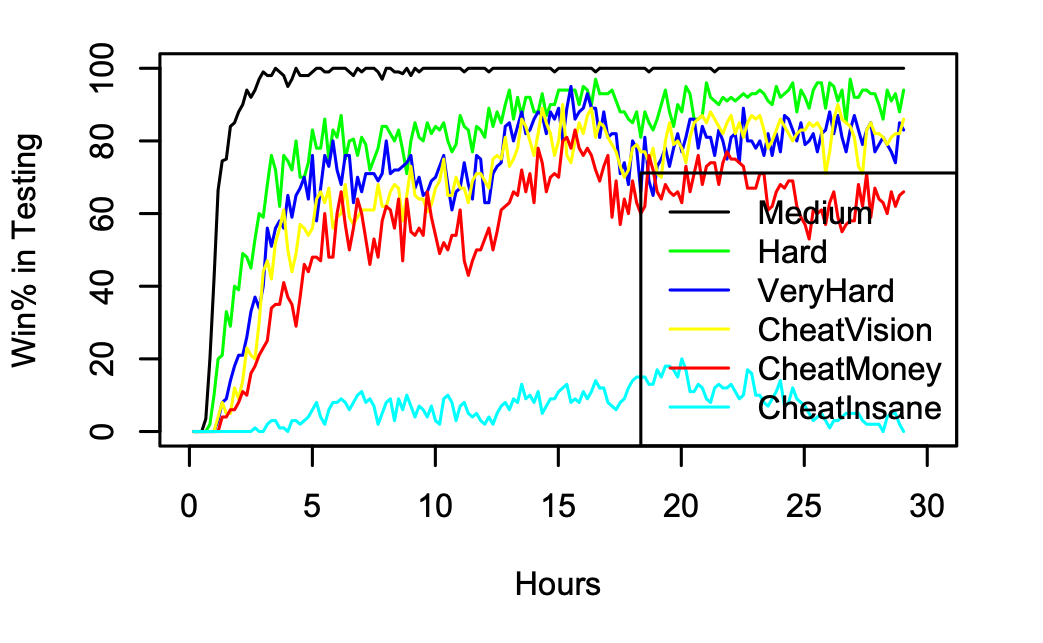}
    \vspace{-2em}
    \caption{Win rates against built-in AI.}
    \label{fig:TstarBot2-result}
\end{figure}

We compose the observation interfaces above and action interfaces of TstarBot2, then train the model from scratch with a MARL algorithm which performs pure ``Self-Play'' with opponents uniformly sampled from all historical agents. We restrict the agents to taking one macro action every 32 frames (i.e.~about 2 seconds). After about $1$ day of training with a single P40 GPU and 2800 CPUs, the deep neural network model can beat the built-in AI from level-1 to level-8 with over $90\%$ winning rate. Figure \ref{fig:TstarBot2-result} shows the learning curve of win-rate against different levels of built-in AI during training.

\subsection{Pommerman}
Pommerman \cite{resnick2018pommerman} is a popular environment based on the game of Bomberman for multi-agent learning. A typical game of Pommerman has 4 agents, each can move and place bomb on the playground. The action space for each agent is a discrete space of 6 actions: \{Idle, Move Up, Move Down, Move Left, Move Right, Place a Bomb\}. The limit of episode length is 800 steps. The game supports free for all (FFA) playing and 2v2 settings. In the FFA mode, each agent's goal is to survive and kill others with bombs; while in 2v2 mode, agents cooperate as a team of 2 and battle against the other team. For the ease of setting up FFA matches as well as 2v2 battles, we provide the following interfaces to be used with Pommerman environment and agents.
\begin{description}
\item[BoardMapObs] Append to observation an image-like feature of agents and items.
\item[AttrObsItf] Append to observation a vector-like agent attributes.
\item[ActMaskObsItf] Append to observation the availability of each action.
\item[RotateItf] Rotate the board (and action) according to agent id.
\end{description}

\begin{figure}[ht]
    \centering
    \includegraphics[width=0.6\textwidth]{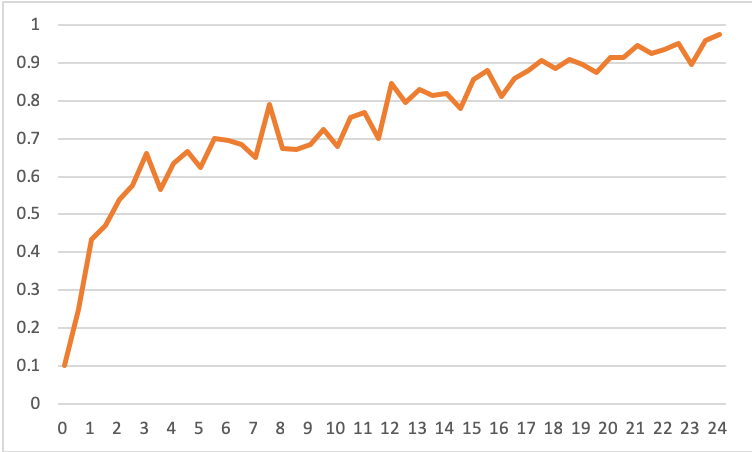}
    \caption{Win rates against rule-based SimpleAgent.}
    \label{fig:pommerman-result}
\end{figure}

Using these interfaces, we train a centralized model controlling two agents as a team in 2v2 mode using pure ``Self-Play'' from scratch with a single P40 GPU learner and 160 actors on 8 remote CPU machines. Figure \ref{fig:pommerman-result} shows the learning curve of win-rate against two rule-based \verb|SimpleAgents| \cite{resnick2018pommerman} during 24 hours' training.

\subsection{ViZDoom}
ViZDoom~\cite{Kempka2016ViZDoom} is an AI research platform based on the classical First Person Shooter game Doom.
The most popular game mode is probably the so-called \emph{Death Match},
where several players join in a maze and fight against each other.
After a fixed time, the match ends 
and all the players are ranked by the FRAG scores 
defined as kills minus suicides.
During the game, 
each player can access various observations, 
including the first-person view screen pixels, 
the corresponding depth-map and segmentation-map (pixel-wise object labels),
the bird-view maze map, etc.
The valid actions include almost all the keyboard-stroke and mouse-control a human player can take,
accounting for moving, turning, jumping, shooting, changing weapon, etc.
ViZDoom can run a game either synchronously or asynchronously,
indicating whether the game core waits until all players' actions are collected or runs in a constant frame rate without waiting.

Since the initial release of ViZDoom, 
several annual AI competitions had been held~\cite{wydmuch2018vizdoom},
attracting the attentions of both academia and industry~\cite{wu2016training,dosovitskiy2016learning,huang2019combo}.
The AI competition intends for encouraging the visual observation based RL, 
hence it constrains the observation a player can use~\cite{wydmuch2018vizdoom}.
Roughly, only the first person view screen pixels are allowed.
Also, the health point, the ammo and the armor are directly provided. These three numbers can be read from the panel showing up in the screen~(Fig.\ref{fig:screenshot}, lower middle).
The competition runs in synchronous mode, 
demanding the real-time decision of the submitted AI.
There is no particular restriction over the action,
except for that always crunching is disallowed,
as it avoids being seen and shot by other players in a too tricky way.
These settings closely emulate a human player's observation and behavior.

Arena provides several interfaces for customizing the observations and actions.
Below we give some examples.
\begin{description}
\item[ScreenAsObs] RGB screen pixels in 2d image.
\item[SegmentationAsObs] Pixel-wise object labels.
\item[DiscreteAction7] Simplify the action space and expose 7 discrete actions, accounting for moving forward, turning left/right, shooting, etc.
\item[PostRuleAction] Add helpful rule-based auto-actions as described in~\cite{wu2016training} that accelerates turning and moving, forces firing suppression, avoid blocking by the explosive bucket, etc.
\item[GameConfigRelated] Such kind of interface changes various game configurations, e.g., the episode length, the synchronous or asynchronous mode, the map loaded, the player's color, etc. 
\end{description}

By combining and building on top of the interfaces, 
the user is able to conveniently test and compare different deep RL methods .
For example,
one can take advantage of the ground-truth segmentation labels.
During training, 
both the \emph{ScreenAsObs} and the \emph{SegmentationAsObs} are taken,
with a neural network architecture that takes as input both the RGB screen pixels and the segmentation map.
During the competition testing, 
only the screen pixels interface \emph{ScreenAsObs} is taken and a separate neural network component is employed to perform the image segmentation.
For another example, 
one can leverage the \emph{PostRuleAction} interface to enhance the agent and accelerate its exploration during RL training.

\begin{figure}[ht]
    \centering
    \includegraphics[width=0.6\textwidth]{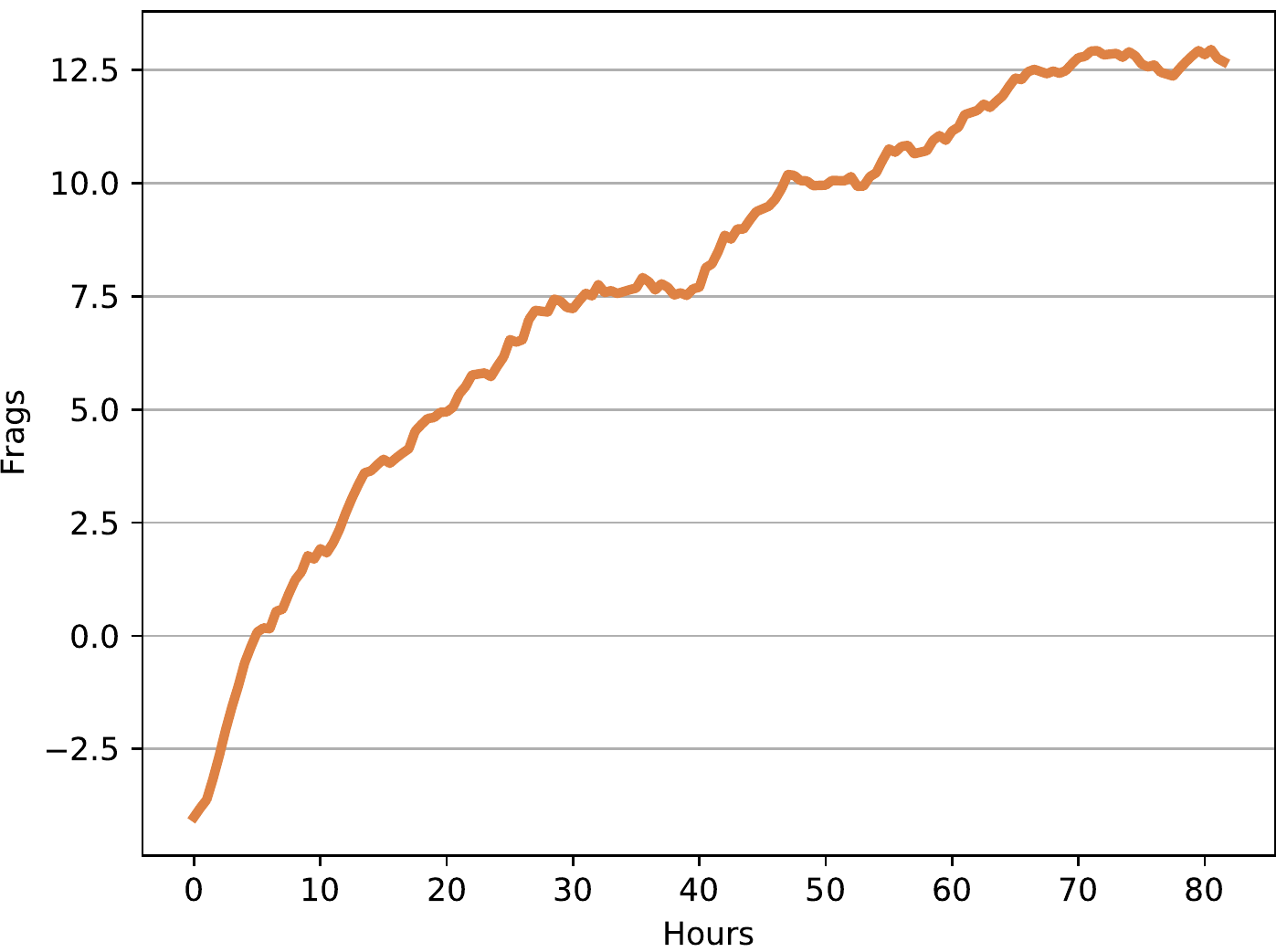}
    \caption{Episodic Frags against built-in AI in ViZDoom Death Match. Horizontal: self-play training hours; Vertical: Frags averaged in 3 episodes in separate testing. }
    \label{fig:vizdoom-result}
\end{figure}

We compose some of the above interfaces and train the model from scratch with a MARL algorithm that performs ``Self-Play'' which is reminiscent to~\cite{bansal2017emergent,jaderberg2018human}.
We employ 4 Learners (each occupying a GPU) and 2000 actors.
Over time,
the neural network model becomes gradually stronger.
When testing it in an 8-player Death Match with 7 built-in Bots unseen during training, 
the model's FRAG score increases reasonably with the training time,
as shown in Fig.~\ref{fig:vizdoom-result}.
It validates the Arena interfaces we provide.

\subsection{Soccer}
The Soccer environment is introduced by the work of \cite{liu2019emergent}. Unlike the previous environments, Soccer is a continuous control environment for multi-agent reinforcement learning and is originally built based on the MuJoCo physics engine~\cite{todorov2012mujoco}. In the game, there are 2 teams and each team has 2 players. 
A player has a single sphere (the body) with 2 fixed arms, and a box head. It has a 3-dimensional action space: accelerate the body forwards/backwards, rotate around the vertical axis of the body with torque, and apply downwards force.
Torque helps the player spin, steer, or kick the football with its arms with more force. 
The contacts between the players are disabled, but the contacts between the players, the pitch and the ball are used.
At each timestep, observation is a 93-dimensional vector, including proprioception (position, velocity, accelerometer information), task (egocentric ball position, velocity and angular velocity, goal and corner positions) and teammate and opponent (orientation, position and velocity) features.
Each soccer match lasts upto 45 seconds, and is terminated when the first team scores. The players choose a new action every 0.05 seconds.

Simply, we describe the interfaces for Soccer environment and agents as follows:
\begin{description}
\item[Dict2Vec] Convert dictionary-style observation to vector representation.
\item[MakeTeam] Group agents to two teams.
\item[ConcatObsAct] Concatenate the observations of multiple agents as a vector and concatenate the actions of agents as a vector.
\end{description}

\begin{figure}[ht]
    \centering
    \includegraphics[width=0.6\textwidth]{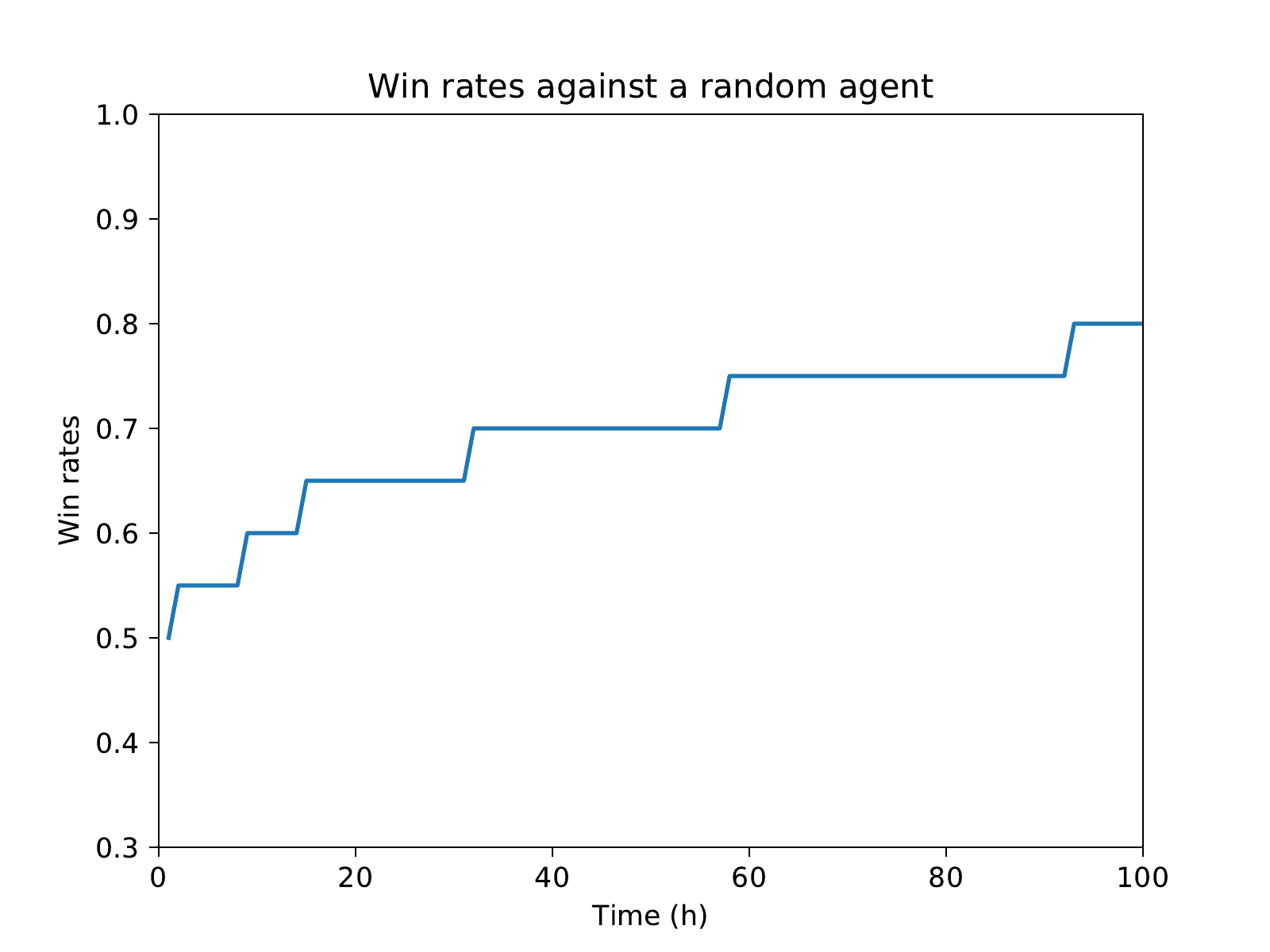}
    \caption{Win rates against a random agent.}
    \label{fig:soccer}
\end{figure}

We use proximal policy optimization (PPO) algorithm to update the policy of our agent based on the platform. We also use self-play training and reward shaping. We consider 4 kinds of rewards: home score, away score, velocity to ball, and velocity from ball to goal. Their weights are $1.0, -1.0, 0.001,0.002$. For evaluation, we consider a simple random agent as an opponent, which is used in~\cite{liu2019emergent}. Our agent wins against a random agent on 80.0\% of the games (a win is worth 1 point and a draw 0.5 points). Figure~\ref{fig:soccer} shows the performance of the best model up to different time points. It shows that the performance increases as training progresses. It does not work better than methods with PBT (Population-based Training) proposed by \cite{liu2019emergent}. However, it works better than other methods without PBT. For example, a feedforward policy with action-value estimator proposed by \cite{liu2019emergent} just achieves 65.0\%.

\end{document}